\documentclass[conference]{IEEEtran}
\IEEEoverridecommandlockouts
\usepackage{cite}
\usepackage{amsmath,amssymb,amsfonts}
\usepackage{algorithmic}
\usepackage{graphicx}
\usepackage{textcomp}
\usepackage{xcolor}
\usepackage{url}
\usepackage{multirow}
\usepackage{graphicx}
\usepackage{booktabs}
\def\BibTeX{{\rm B\kern-.05em{\sc i\kern-.025em b}\kern-.08em
  T\kern-.1667em\lower.7ex\hbox{E}\kern-.125emX}}

\begin{document}

\title{Comparison of Machine Learning Models in Food
  Authentication Studies}

\author{\IEEEauthorblockN{Manokamna Singh}
  \IEEEauthorblockA{
    \textit{CeADAR, School Of Computer Science, UCD}\\
    Dublin, Ireland\\
    manokamna.singh@gmail.com}
  \and
  \and
  \IEEEauthorblockN{ Dr. Katarina Domijan}
  \IEEEauthorblockA{\textit{Department of Mathematics and Statistics, Hamilton Institute} \\
    \textit{Maynooth University}\\
    Maynooth, Co Kildare, Ireland \\
    katarina.domijan@mu.ie}
}

\maketitle 

  \IEEEpubid{978-1-7281-2800-9/19/\$31.00\copyright\ 2019 IEEE}

\begin{abstract}
The underlying objective of food authentication studies is to determine whether unknown food samples have been correctly labelled. In this paper we study three near infrared (NIR) spectroscopic datasets from food samples of different types: meat samples (labelled by species), olive oil samples (labelled by their geographic origin) and honey samples (labelled as pure or adulterated by different adulterants). We apply and compare a large number of classification, dimension reduction and variable selection approaches to these datasets. NIR data pose specific challenges to classification and variable selection: the datasets are high - dimensional where the number of cases ($n$) $<<$ number of features ($p$) and the recorded features are highly serially correlated. In this paper we carry out comparative analysis of different approaches and find that partial least squares, a classic tool employed for these types of data, outperforms all the other approaches considered. 
\end{abstract}
\IEEEpubidadjcol


\section{\textbf{Introduction}}

In this paper we analyse three infrared (NIR) spectroscopic datasets from food authentification studies. The aim of such studies is to determine whether the food samples have been correctly labelled. This translates to a multi-category supervised classification problem, where a classifier is trained on a set of samples and the known and predicted labels are compared. For those samples that are determined to be potentially inacurately labelled, further testing can be used.  The  NIR datasets are made up of features of transflectance spectra recorded over a range of wavelengths. The adjacent features are highly correlated and the datasets are of high dimension where the number of features ($p$) exceeds many-fold the number of cases ($n$), $p  <<  n$. Common approaches to classifying NIR spectroscopic data involve Partial Least Squares (PLS) or Linear Discriminant Analysis (LDA) \cite{ref31ldafisher} with a Principal Components Analysis (PCA) pre-processing step. In this work, we present a comparative analysis of a range of machine learning classification techniques for three NIR datasets, namely LDA, Quadratic Discriminant Analysis (QDA) \cite{ref16lda1}, k-Nearest Neighbours (kNN) \cite{ref16lda1,ref18mass}, Decision Trees \cite{c45}, Logit Boost \cite{Dettling03boostingfor}, Random Forest\cite{ref37rf}, Support Vector Machine \cite{vapnik98}, Bayesian Kernel Projection Classifier (BKPC)\cite{domijan09} and Partial Least Squares \cite{ref44pls}. We combine them with a range of dimension reduction and variable selection approaches as preprocessing steps, namely: Principal Component Analysis (PCA) \cite{Reference8}, Functional Principal Component Analysis (FPCA) \cite{Ref11fpca}, Marginal Relevance (MR) \cite{ref14dudoit} and Genetic Algorithm (GA) \cite{ref50ga}. Our conclusion is that none of the approaches considered were able to outperform PLS in any of the three datasets considered. 
\IEEEpubidadjcol

\section{\textbf{Methodology}}

\subsection{\textbf{Near Infrared Data}}

\begin{figure}[htbp]
\centerline{\includegraphics[width=80mm]{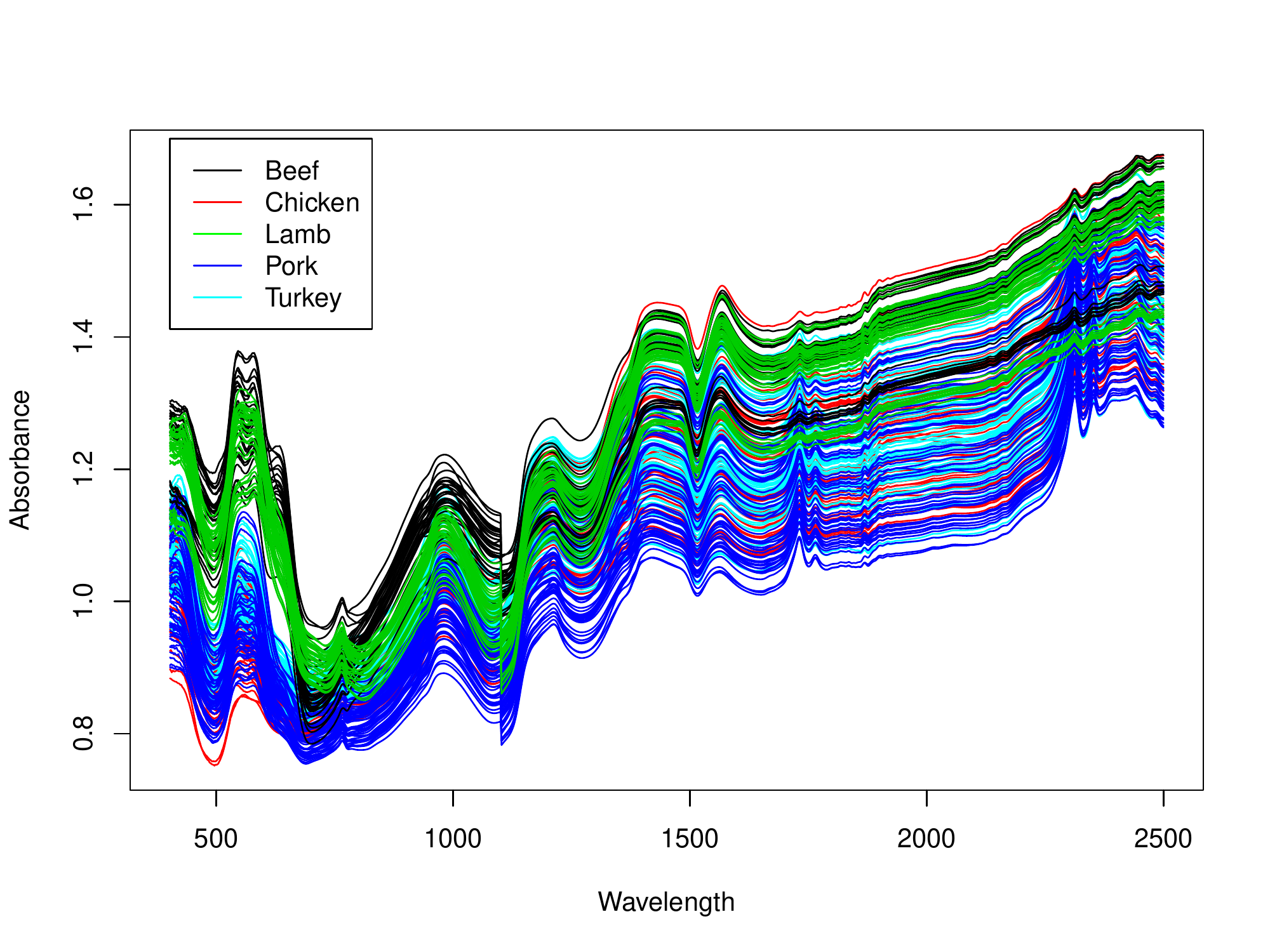}}
\caption{NIR spectra of raw homogenised meat samples.}
\label{fig:meat}
\end{figure}

\begin{figure}[htbp]
\centerline{\includegraphics[width=80mm]{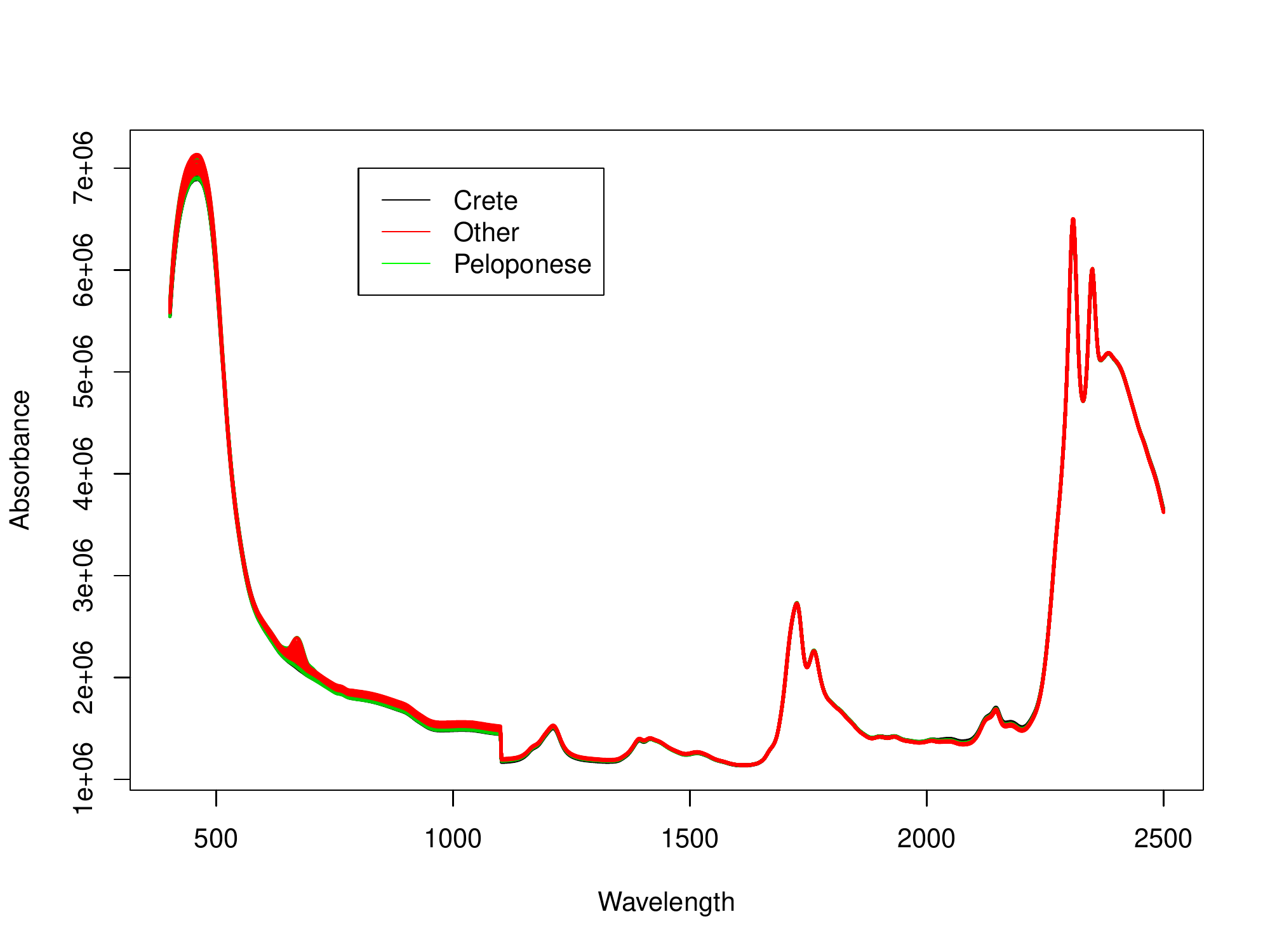}}
\caption{NIR spectra of olive oils.}
\label{fig:olive}
\end{figure}

\begin{figure}[htbp]
\centerline{\includegraphics[width=80mm]{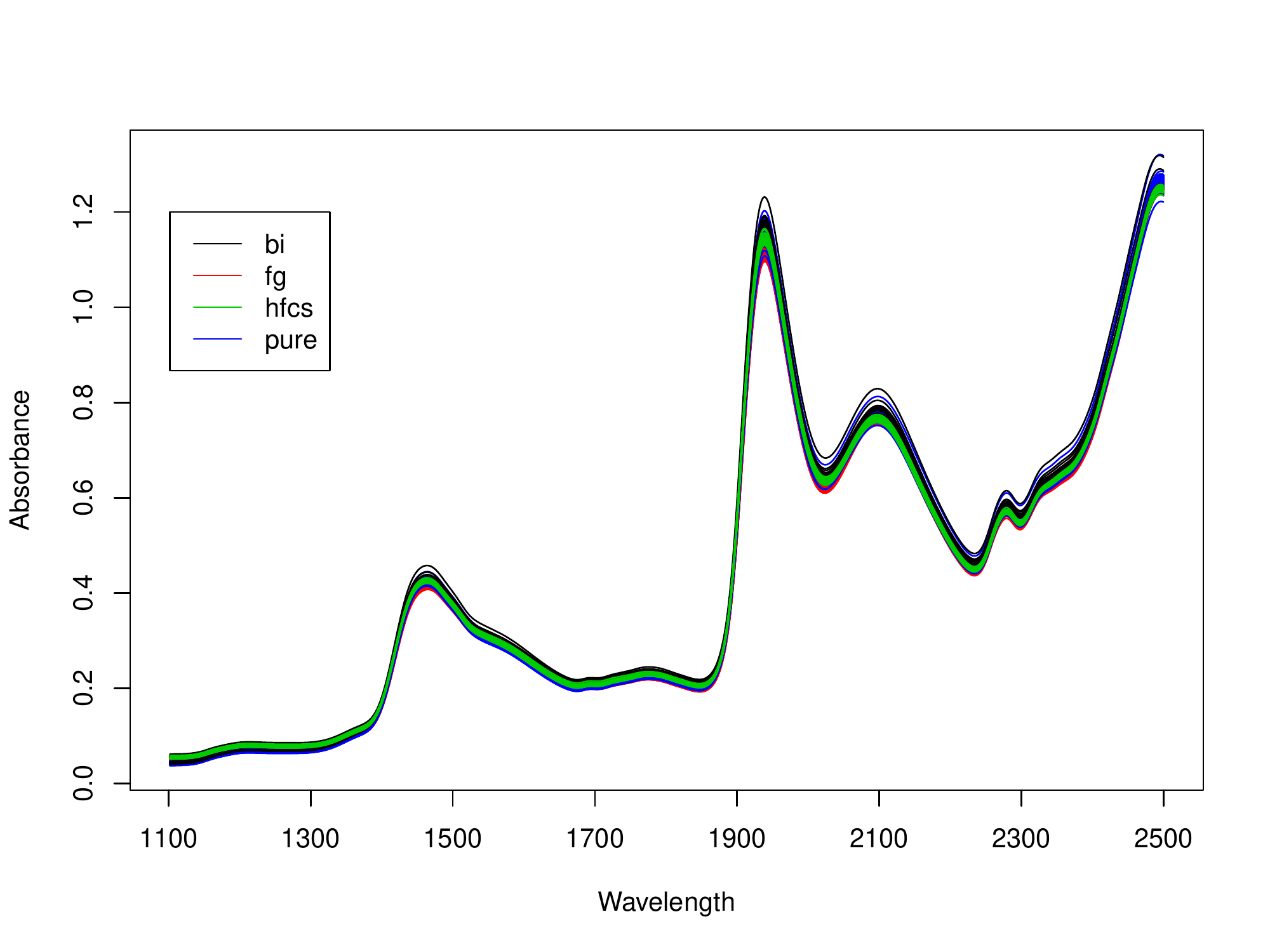}}
\caption{NIR spectra of pure and adulterated honey samples.}
\label{fig:honey}
\end{figure}

The three NIR datasets presented in this paper are meat \cite{ref45meatdata}, honey \cite{ref47honeydata} and olive oil \cite{ref46oildata}. All have been previously  analysed using model based clustering approaches \cite{ref53discri,Ref1toher, Ref2murphy}. The meat dataset comprises $1051$ features recorded for $231$ samples. Each sample is labelled by one of $k = 5$ classes:  beef, chicken, lamb, pork, turkey, see Fig \ref{fig:meat}.  Olive oil data comprises $1051$ features over $65$ samples with 3 labels denoting the geographic origin of oil (Crete, Peloponnese, Other), see Fig \ref{fig:olive}. Honey data contains spectra of $478$ samples of pure and adulterated samples recorded over $700$ wavelengths, see Fig \ref{fig:honey} The adulterated honey samples are further labelled by the the type of adulteration: fully-inverted beet syrup (bi), fructose:glucose mixtures (fg) and high fructose corn syrup (hfcs), thus there are 4 classes in total. The datasets are summarised in Table \ref{NIRdata}.

\begin{table}[ht]
\caption{NIR datasets summarised by the number of samples ($n$), number of features ($p$) and the number of classes ($k$).}
\begin{center}
\begin{tabular}{l|c|c|c}

   & $n$ & $p$ & $k$  \\
   \hline
Meat   & 231 & 1050  & 5    \\
Olive oil   & 65 & 1050  & 3   \\ 
Honey    & 478 & 700   & 4  \\ 

\end{tabular}\label{NIRdata}
\end{center}
\end{table}

\subsection{\textbf{Data Splitting}}

All three datasets were randomly split into training (50 \% and test (50 \%) sets. The random sampling was done within each class such that it preserved the overall class distribution of data. All the dimension reduction, variable selection and classification algorithms were trained on the training sets and tested on the testing sets over 100 such random splits of the data. 

\subsection{\textbf{Dimension Reduction and Variable Selection}}

High dimensional data, in particular when $p >> n$, pose a convergence problem for a number of classification techniques. Multi-collinearity arises since any variable can be written as a linear combination of all the other variables in the model thus it is impossible to find the best subset of variables useful for prediction. In addition, features in the NIR datasets are highly serially correlated as absorbances recorded at adjacent wavelenghts are very similar as can be seen from Figures \ref{fig:meat}- \ref{fig:honey} which exacerbates the problem.  A different training set is likely to identify a different subset of features as useful for prediction. 

Kernel based algorithms, such as Support Vector Machines (SVMs) \cite{vapnik98}, evade this problem as they condense the data into a distance matrix between the samples whose dimensionality does not depend on $p$. Whereas high collinearity is not a problem for kernel methods, datasets where only a small subset of features contain information on class discrimination require some feature selection preprocessing as all features contribute equally to the kernel and the irrelevant features are likely to obscure the signal and lead to degraded prediction performance. 

In this paper we utilise and compare a range of approaches to reducing the dimensionality of the training set.

\subsubsection{\textbf{Principal Component Analysis}} 

\begin{figure}[htbp]
\centerline{\includegraphics[width=80mm]{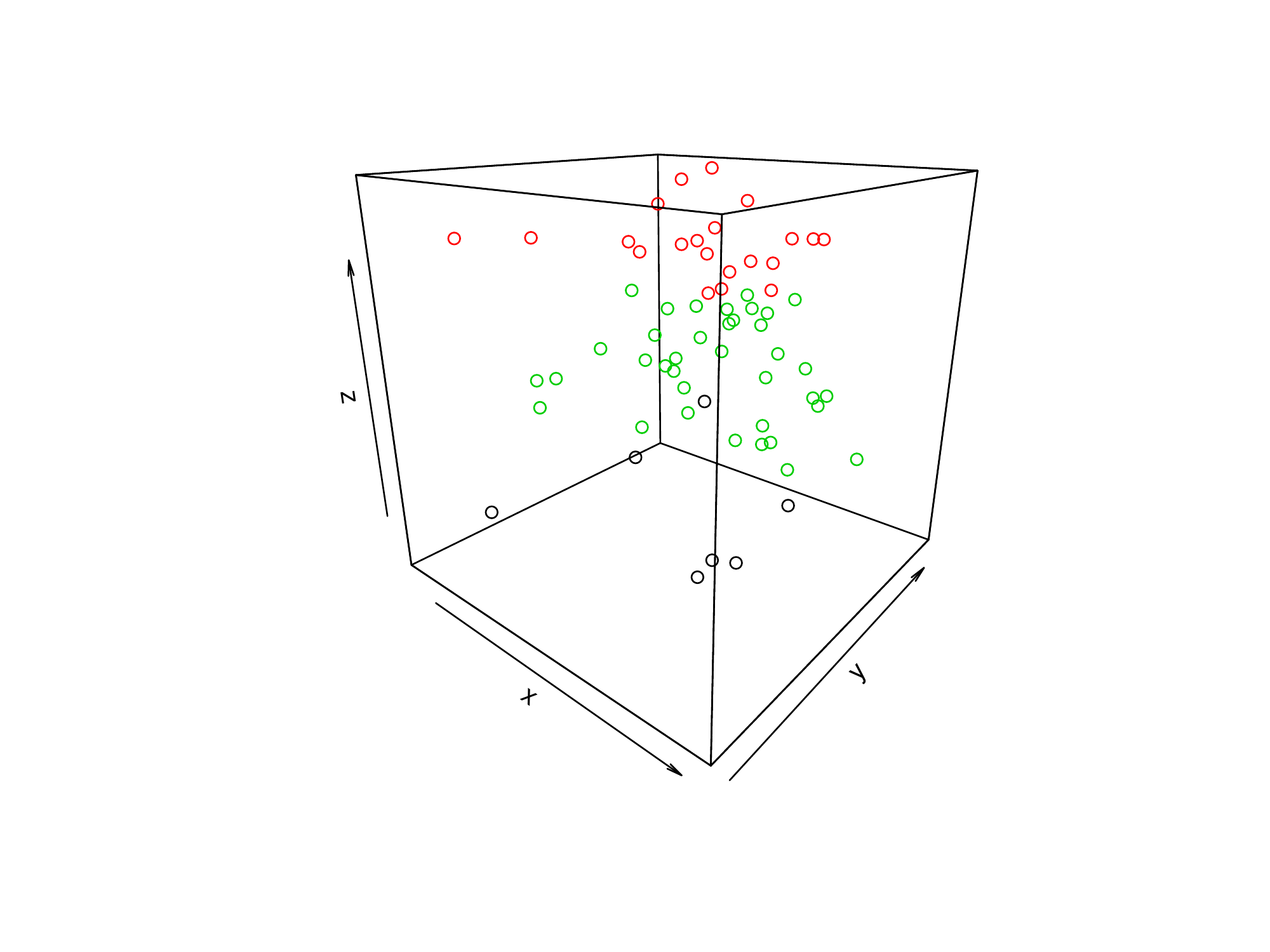}}
\caption{First three PCs of NIR spectra of olive oils.}
\label{fig:olivePCs}
\end{figure}

Principal Component Analysis (PCA) \cite{Reference8} obtains lower dimensional projections of the data in the feature space. The new features (PCs) are uncorrelated linear combinations of the original features and represent directions in the observation space along which the data have the highest variability. In this application, PCA was carried out on the training set of each random split of the data. The number of PCs required to explain 99\% of the variability was 4.5 for meat data, 12.32 for oil data and 5.47 for honey data, averaged over 100 random splits, thus significantly reducing the dimensionality of the input feature space from 1050 or 700 features, see Figure \ref{fig:olivePCs}.  
PCA is a dimension reduction method that works well at reducing dimensionality for highly correlated datasets and has a good track record as a preporocessing tool in NIR datasets. It is an unsupervised method, since the class information is not utilised in the dimension reduction. 

\subsubsection{\textbf{Functional Principal Component Analysis (FPCA)}}

Functional Principal Component Analysis (FPCA) \cite{Ref11fpca} is the PCA analogue for functional data. In this framework, absorbance values for each sample are viewed as continuous functions of wavelengths sampled at particular intervals and subject to observational noise. The techniques of functional data analysis are used to carry out PCA directly on these functions. The functional principal component scores are then used as feature inputs to the classification algorithms. Note that whereas PCA assumes no particular ordering of the features in the data space, the absorbances of NIR data are recorded over an ordered range of wavelengths and FPCA is able to utilise this information. Like PCA, it takes all the available features as input and does not use class information in the training set to obtain the lower dimensional representation of te data. The number of FPC scores was set at 4 for all three datasets.


\subsubsection{\textbf{Marginal Relevance (MR)}}

The above described approaches (PCA and FPCA) find lower dimensional representation of the data by taking all the features as input. In contrast, we also considered to approaches to selecting a small subset of wavelengths for classification. Marginal Relevance (MR) \cite{ref14dudoit} is a criterion that ranks each feature in order of their capability to discriminate between the classes. The MR score for each feature is the ratio of the between-class to within-class sum of squares, an idea underpins many statistical methodologies and is a frequently used filtering technique to effectively screen out large numbers of spurious features. A serious drawback of this approach is that each feature is considered independently of others so the highest-ranked features can be correlated and do not necessarily form the optimal subset for the purposes of classification. Also it is not able to select features that contain information on class discrimination jointly with others, but not marginally. In this application, in each dataset we select 10 features with highest MR score as input to classification algorithms.

\subsubsection{\textbf{Genetic Algorithm}}

Variable selection for high-dimensional datasets is a difficult problem as the number of possible configurations of the feature set where individual wavelengths are either included or excluded is $2^{p - 1}$. As the number of features $p$ gets large, the model space approaches infinity and selecting a globally optimal subset of features by traversing the entire model space in finite time is an impossible task. Furthermore, the space of possible models is discrete and is likely to exhibit many undesirable properties such as multimodality.

In this paper, we employ Genetic Algorithm (GA) \cite{ref50ga}, a stochastic search technique to explore the space of all possible feature configurations. The algorithm works by analogy to natural evolution. It starts by creating a population of randomly selected variable configurations  and evaluates their fitness using criteria such as classification accuracy. The feature configurations propagate themselves through subsequent generations using operators such as mutation, crossover and fitness based selection. In the implemenation used, the algorithm terminates after a fixed number of generations and reports a list of features most commonly selected in all populations. In this paper we take 5 top features selected by GA as inputs for classification algorithms.

\subsection{\textbf{Classification}}

\textbf{Partial Least Squares} (PLS) is a classic approach to NIR datasets, specifically developed for regression and classification problems in chememotric applications where a large number of highly correlated features is recorded for each observation. Like PCA, it seeks to find a lower dimensional representation of the data in the form of a smaller set of linear combinations of the original features. However,  PLS is a supervised alternative, where the response is taken into account in determining the direction of the projection. This is obtained through iterative procedures and a number of algorithm variants exist. Subsequently, linear regression or classification are carried out using a subset of PLS scores as inputs.

\textbf{Linear Discriminant Analysis} (LDA) is multi-category classification method. It fits a $p$-dimensional multivariate normal density to observations from each class. The densities have different estimated means, but are constrained to have the same covariance structure. New observations are allocated to the class with the highest estimated density value for that observation (i.e. minimum Mahalanobis distance to the mean). This leads to linear discrimination boundaries in the feature space between the classes. \textbf{Quadratic Discriminant Analysis} (QDA) is an extension where the restriction on the covariance structure is relaxed, which leads to a nonlinear classification boundary. Neither method is appropriate for $p>>n$ data and both require preprocessing steps to reduce dimensionality of the data. QDA requires a larger number of parameters to be estimated than LDA.

\textbf{k-nearest neighbors} (kNN)  algorithm is a non-parametric classification method. It classifies each observation based on the majority class of its $k$ nearest (in Euclidean distance) neighbours. It is more flexible than LDA and QDA as it does not make distributional assumptions about the data and allows for highly nonlinear decision boundaries. Tuning parameter $k$ is estimated using cross-validation of the training set.

\textbf{Decision trees} (DCT) for classification recursively partition the feature space into rectangular subregions where the predicted class is the most common occurring class. At each iteration, a tree algorithm searches through all of the possible split-points of all the features to find a partition which minimizes a specified criterion. In  classification problems this criterion can be the classification error rate, but is often taken to be a measure of region impurity as defined by the Gini index or cross-entropy. Like kNN, this is a non-parametric method that allows for nonlinear, but less flexible, decision boundaries. The set of split decisions can be mapped to a tree diagram, which has the advantage of making the final classifier interpretable.  

\textbf{Random Forrests} (RF) extend the above approach by growing a number of decision trees on bootstrapped samples of the training set. In order to decorrelate the trees, only a small subset of randomly selected features is considered when searching for the optimal split-points. Finally a single consensus prediction is obtained from all the trees using majority vote. The total decrease in the Gini index from splitting on a feature, averaged over all trees, can be taken as an estimate of that feature's importance. Random forests are known to outperform the decision trees in regression and classification tasks, however, the interpretability of the trees is lost.

\textbf{LogitBoost} (LB) is a modified version of boosted decision trees for classification. Boosting is an alternative approach to bagging for improving the predictions resulting from decision trees. Whereas RFs employ bagging strategy of creating multiple copies of the original training dataset using the bootstrap, boosting grows the trees sequentially: using information from previously grown trees (weak learners). The LB classifier uses decision stumps (one node decision trees) as weak learners and utilises a feature preselection method related to MR, where the score is equivalent to test statistic of Wilcoxon’s two sample test. The method was developed for $p>>n$ gene expression data.

\textbf{Support Vector Machine} (SVM) is a kernel extension of a binary linear classifier that constructs a hyperplane to separate two classes. The hyperplane is chosen so that the smallest perpendicular distance of the training data to the hyperplane (margin) is maximized. A tuning parameter (cost) controls the number of observations that are allowed to violate the margin of the hyperplane. The kernel trick is a general technique that implicitly maps the observations to a higher-dimensional feature space where the classes are linearly seperable. For linear classifiers, this has the effect of fitting nonlinear decision boundaries in the original feature space. The shape of the decision boundary is determined by the choice of the kernel and its parameterization.

\textbf{Bayesian Kernel Projection Classifier} (BKPC) is a multi-category kernel classification method. Like SVM, it uses the kernel trick to implicitly map the data into a high dimensional feature space where the classes are thought to be
linearly separable. However, BKPC performs the classification of the projections of the data to the principal axes of the feature space, which is a related approach to PCA. The model is fitted in the Bayesian framework, thus probability distributions of prediction can be obtained for new data points.

\section{\textbf{Results}}\label{result}

The analysis of the data was done in R, a free software environment for statistical computing and graphics \cite{R17}.  

Marginal relevance (MR) criterion  is implemented in R package BKPC \cite{Ref9bkpc}. For all datasets, we took the top 10 highest scoring features as inputs into the classification algorithms. R has a number of implementations for FPCA, but most lack a predict method. The implementation used in this paper is in the fdapace \cite{fpcaR} package.  GA is implemented in GA \cite{Reference7} package. We used caret \cite{Ref10caret} as a front end to GA for variable selection. This implementation takes RF model and 10 fold cross validation to assess the fitness of each feature configuration. The algorithm was ran for 100 generations with population size of 50. Crossover and mutation probability were set at 0.8 and 0.1 respectively.

LDA and QDA are implemented in MASS \cite{ref18mass} package and require no parameter tuning. PLS is implemented in package pls \cite{ref23pls}. DCT implementation was the C4.5 algorithm in package RWeka \cite{ref24dct}.
kNN, RF, LB, SVM and BKPC are implemented in packages class \cite{ref18mass}, randomForest \cite{ref22rf}, caTools \cite{ref21lgr}, e1071 \cite{ref43svm} and BKPC\cite{Ref9bkpc} respectively. Gaussian kernel was used for the SVM with the bandwidth parameter value set at an empirical estimate suggested by \cite{caputo}.

All the feature selection and classification techniques were trained on the training sets of the data and the prediction was obtained using the test set only. Methods that required parameter tuning used further cross-validation of training set over a grid of possible values. The process was repeated for the 100 random splits of the data except for GA which was carried out over only one split of the data due to computational cost of the implementation. Average classification accuracy (ACC) over 100 random splits and the associated standard deviations (SD)  are reported in the Table \ref{compareresult}. LDA, QDA, kNN and DCT were applied after preprocessing with PCA, FPCA, MR and GA. LB, RF and SVM were trained after the four preprocessing steps and on the entire data without any preprocessing for dimension reduction. PLS was applied without any preprocessing. The number of PLS scores was set at 15, 10 and 15 for meat, oil and honey data respectively. BKPC was trained without any preprocessing and on the top 10 features selected by MR. The number of principal axes was taken to be 10, 20 and 20 for the meat, olive oil and honey data respectively.

Since the training is done on only one split of the data, GA results have no associated SD.  No resultsa are reported for QDA with PCA and MR are preprocessing steps for the olive oil data as the method did not converge due to the small sample size: the training sets contained only $32$ observations over three classes. 

The comparative analysis suggests that PLS consistently performs better for NIR data with respect to the range of machine learning algorithms considered. It also performs competitively with the model based clustering and variable selection approaches taken by \cite{Ref1toher} and \cite{Ref2murphy} who take a similar cross-validatory approach and report accuracy of $93.5$\%, $93$\% and $94$\% for meat, olive oil and honey datasets respectively.  

The confusion matrices obtained from the test data in the 100th random split of PLS with the meat, olive oil and honey data are shown in Tables \ref{meatres}, \ref{oilres} and \ref{honeyres}. We observe that in the meat dataset almost all miss-classified samples come from the poultry groups (chicken/turkey) and the hardest classes to separate in the honey data are pure honey and the fructose:glucose mixtures (10 miss-classified observations).

\begin{table}
\caption{Classification results for meat, honey and oil data}

\begin{center}
\begin{tabular}{|l|l|l|l|l|l|l|}
\hline
\multicolumn{1}{|c|}{} & \multicolumn{2}{c|}{Meat} & \multicolumn{2}{c|}{Olive Oil} & \multicolumn{2}{c|}{Honey} \\ 
\hline
\hline
Model &  ACC &   SD  &  ACC &  SD  &  ACC  & SD    \\ 
\hline
\hline
 LDA PCA & 81 & 0.03 & \textbf{91} & 0.04& 80 & 0.03  \\ 
 LDA FPCA & 68 & 0.03 & 53 & 0.06 & 70 & 0.02 \\ 
 LDA MR & 79 & 0.03 & 81 & 0.07 & 86 & 0.03  \\ 
 LDA GA & 88 & NA & 72 & NA & 81 & NA\\
\hline
\hline
 QDA PCA & 82 & 0.03 & NA & NA& 89 & 0.02 \\ 
 QDA FPCA & 71 & 0.03 & 63 & (0) & 78 & 0.02\\
 QDA MR & 70 & 0.04 & NA & NA& 89 & 0.03 \\ 
 QDA GA & 87 & NA & 56 & NA & 91 & NA\\
\hline
\hline
 kNN PCA & 71 & 0.04 & 76 & 0.07 & 87 & 0.02\\
 kNN FPCA & 77 & 0.04& 64 & 0.07 & 81 & 0.02  \\
 kNN MR & 62 & 0.03 & 60 & 0.07  & 76 & 0.05\\ 
 kNN GA & 79 & NA & 50 & NA & 85 & NA\\
 \hline
 \hline
 DCT PCA & 72 & 0.04 & 70 & 0.09 & 86 & 0.03\\ 
 DCT FPCA & 84 & 0.04 & 70 & 0.08 & 86 & 0.03 \\
 DCT MR & 60 & 0.03 & 60 & 0.07 & 76 & 0.05 \\ 
 DCT GA & 83 & NA & 50 & NA & 84 & NA\\
\hline
\hline
 LB  & 84 & 0.04 & 80 & 0.07 & 89 & 0.02 \\ 
 LB PCA & 80 & 0.04 & 83 & 0.07 & 89 & 0.02 \\
 LB FPCA & 86 & 0.04 & 76 & 0.07 & 88 & 0.02\\ 
 LB MR & 64 & 0.04 & 61 & 0.07 & 78 & 0.05\\ 
 LB GA & 84 &NA & 49 & NA & 85 & NA\\
\hline
\hline
 RF  & 81 & 0.05 & 79 & 0.06  & 89 & 0.02 \\ 
 RF PCA & 77 & 0.03 & 80 & 0.09  & 90 & 0.02 \\ 
 RF FPCA & 82 & 0.04 & 77 & 0.07  & 86 & 0.02\\ 
 RF MR & 60 & 0.03 & 61 & 0.08 & 73 & 0.06 \\ 
 RF GA & 87 &NA & 53 & NA  & 89 & NA\\

\hline
\hline
SVM & 77 & 0.04 & 80 & 0.06 & 89 & 0.01\\ 
SVM PCA & 70 & 0.05 & 86 & 0.06 & 91 & 0.02\\ 
SVM FPCA & 78 & 0.05 & 67 & 0.07 & 80 & 0.03\\
SVM MR & 63 & 0.03 & 59 & 0.08 & 77 & 0.06\\ 
SVM GA & 83 & NA & 56 & NA & 92 & NA\\
\hline
\hline
BKPC & 76 & 0.04 & 72 & 0.07 & 81 & 0.05\\ 
BKPC MR & 71 & 0.05 & 63 & 0.07 & 81 & 0.05\\ 
\hline 
\hline
PLS & \textbf{94} & 0.02 & \textbf{90} & 0.05 & \textbf{95} &  0.01\\
\hline
\end{tabular}
\label{compareresult}
\end{center}
\end{table}

\begin{table}[ht]
\caption{PLS Result: Meat}
\begin{center}
\begin{tabular}{cc|ccccc}
\multicolumn{1}{c}{} & \multicolumn{6}{c}{Reference} \\
&   & Beef & Chicken & Lamb & Pork & Turkey \\ 
\cline{2-7}
\multirow{5}{*}{\rotatebox[origin=c]{90}{Prediction}}
& Beef   & 15 & 0  & 1   & 0 & 0    \\
& Chicken    & 0 & 23  & 0   & 0  & 4   \\ 
& Lamb    & 0 & 0   & 17   & 0 & 0  \\ 
& Pork    & 0 & 0   & 0   & 27 & 0  \\ 
& Turkey    & 0 & 1   & 0   & 0 & 26   \\ 
\end{tabular}
\label{meatres}
\end{center}
\end{table}

\begin{table}[ht]
\caption{PLS Result: Oil}
\begin{center}
\begin{tabular}{cc|ccc}
\multicolumn{1}{c}{} & \multicolumn{4}{c}{Reference} \\
&   & Crete & Other & Peloponese \\ 
\cline{2-5}
\multirow{4}{*}{\rotatebox[origin=c]{90}{Prediction}}
& Crete   & 7 & 0  & 2     \\
& Other    & 0 & 9  & 0    \\ 
& Peloponese    & 0 & 0   & 14   \\ 
\end{tabular}
\label{oilres}
\end{center}
\end{table}

\begin{table}[ht]
\caption{PLS Result: Honey}
\begin{center}
\begin{tabular}{cc|cccc}
\multicolumn{1}{c}{} & \multicolumn{5}{c}{Reference} \\
&   & bi & fg & hfcs & pure \\ 
\cline{2-6}
\multirow{4}{*}{\rotatebox[origin=c]{90}{Prediction}}
& bi   & 28 & 0  & 0   & 0     \\
& fg    & 0 & 107  & 0   & 5   \\ 
& hfcs    & 1 & 1   & 18   & 0 \\ 
& Pure   & 0 & 5   & 0   & 73 \\ 
\end{tabular}\label{honeyres}
\end{center}
\end{table}

\section{\textbf{Conclusion}}

Comparative analyses of algorithms have been used in many different application domains, see for example \cite{SOLA}. The success of potential approaches is domain specific and will depend on the structure of the data at hand. NIR data is specific as these high-dimensional datasets have highly serially correlated features. In this study we utilised a range of different approaches to dimension reduction, variable selection and classification with respect to inducing nonlinearity in the decision boundaries and model-fitting. With the exception of PLS, none of the approaches convincingly outperformed the others. For the olive oil data, more complex approaches to variable selection and classification did significantly worse than LDA (a linear classifier) with the PCA (unsupervised) preprocessing step. This is likely to be due to overfitting as the samplem size in this dataset is extremly small ($n = 32$ in the training set for a 3 class problem with $p = 1050$ features).
We find that PLS, a method classically employed for spectroscopic data yielded excellent classification rates and consistently outperformed all other approaches considered including in comparison with other recently published approaches (not replicated in this paper).

\bibliographystyle{IEEEtran}
\bibliography{mybibfile}

\end{document}